\documentclass[10pt,twocolumn,letterpaper]{article}

\usepackage{cvpr}              

\usepackage{graphicx}
\usepackage{amsmath}
\usepackage{amssymb}
\usepackage{booktabs}
\usepackage{algorithm}
\usepackage{algpseudocode}

\usepackage[accsupp]{axessibility}

\usepackage[pagebackref,breaklinks,colorlinks]{hyperref}

\usepackage[capitalize]{cleveref}
\crefname{section}{Sec.}{Secs.}
\Crefname{section}{Section}{Sections}
\Crefname{table}{Table}{Tables}
\crefname{table}{Tab.}{Tabs.}

\def\cvprPaperID{8} 
\def\confName{CVPR}
\def\confYear{2023}

\begin{document}

\title{DynaShare: Task and Instance Conditioned Parameter Sharing for Multi-Task Learning}

\author{
Elahe Rahimian$^1$, Golara Javadi$^2$, Frederick Tung$^1$, Gabriel  Oliveira$^1$ \\
$^1$Borealis AI, $^2$University of British Columbia\\
{\tt\footnotesize \{elahe.najafabadi, frederick.tung, gabriel.oliveira\}@borealisai.com, javadi.gol@gmail.com}
}
\maketitle

\begin{abstract}
Multi-task networks rely on effective parameter sharing to achieve robust generalization across tasks. 
In this paper, we present a novel parameter sharing method for multi-task learning that conditions parameter sharing on both the task and the intermediate feature representations at inference time. In contrast to traditional parameter sharing approaches, which fix or learn a deterministic sharing pattern during training and apply the same pattern to all examples during inference, we propose to dynamically decide which parts of the network to activate based on both the task and the input instance.
Our approach learns a hierarchical gating policy consisting of a task-specific policy for coarse layer selection and gating units for individual input instances, which work together to determine the execution path at inference time. 
Experiments on the NYU v2, Cityscapes and MIMIC-III datasets demonstrate the potential of the proposed approach and its applicability across problem domains.
\end{abstract}

\section{Introduction}
\label{sec:intro}

Multi-task learning focuses on adapting knowledge across multiple related tasks and optimizes a single model to perform all tasks simultaneously. Multi-task networks rely on effective parameter sharing (weight sharing) to achieve robust generalization across tasks. Traditionally, multi-task networks have employed hard or soft parameter sharing strategies. Networks that use hard parameter sharing are composed of a shared backbone of initial layers, followed by a separate branch for each task. The shared backbone learns generic representations, and the dedicated branches learn task-specific representations. Architectures that use soft parameter sharing are composed of multiple task-specific backbones, and parameters are linked across backbones by regularization or fusion techniques. 

In this paper, we propose a novel parameter sharing method for multi-task architectures. In contrast to traditional parameter sharing approaches, which fix or learn a deterministic sharing pattern during training and apply the same sharing pattern uniformly to all examples during inference, we propose to condition parameter sharing on both the task and the intermediate feature representations at inference time. The idea is for the multi-task network to dynamically decide which parts of the network to activate based on both the task and the input instance.

\begin{figure*}[t]
\begin{center}
\includegraphics[width=.99\linewidth]{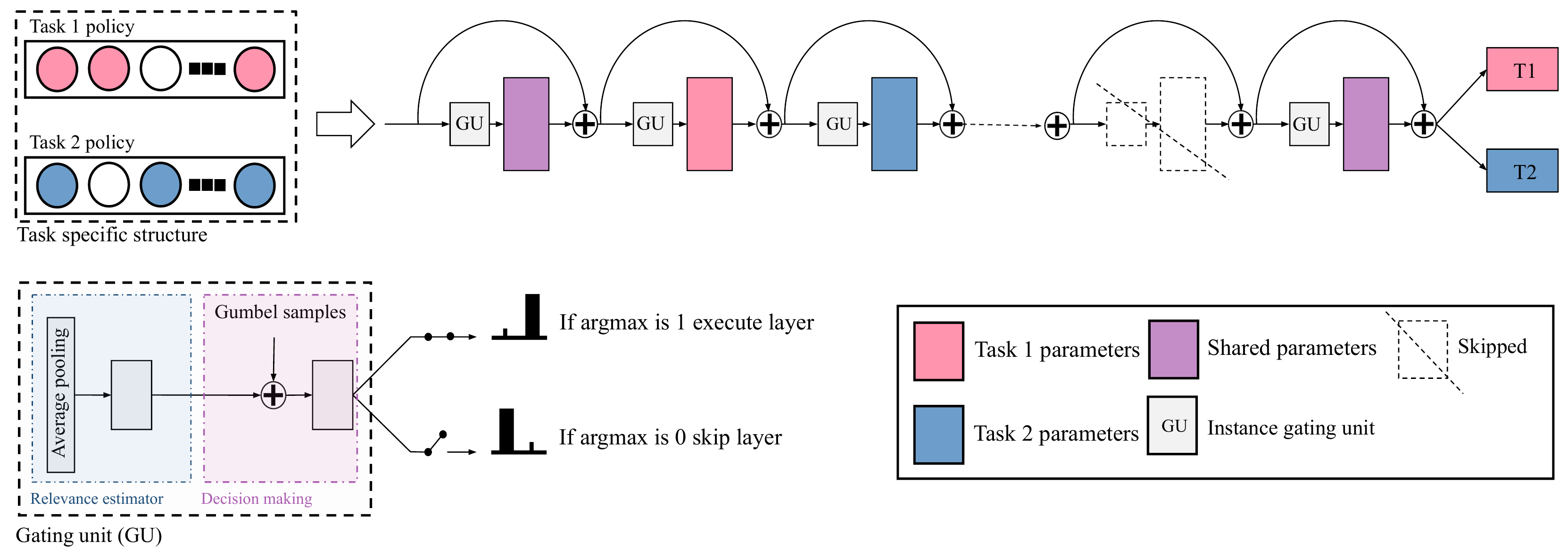}
\end{center}
\caption{An overview of task and instance conditioned parameter sharing for multi-task learning. DynaShare conditions the execution of layers on both the task and the intermediate feature representations.
At inference time, the selected layers for tasks is fixed and the network operates dynamically based on the input instance: the forward path is modified based on the intermediate representations. The gating unit for instance decision making consists of two parts: estimating relevance of an input to the layer, and making the decision for executing the layer based on the relevance.}
\label{fig:workflow}
\end{figure*}

The overall workflow of DynaShare is illustrated in Figure~\ref{fig:workflow}. DynaShare learns a hierarchical gating policy consisting of a task-specific policy for coarse layer selection and gating units for individual input instances, which work together to determine the execution path at inference time.  While dynamic networks are often employed for computational efficiency reasons (e.g. to reduce the inference time footprint with respect to parameters), the primary motivation in this work is to leverage task and instance conditioning to boost the weight sharing flexibility of a multi-task network, with the end goal of enabling better generalization across the multiple tasks.

The contributions of this paper are as follows:
\begin{itemize}
  \item We propose a novel and fully differentiable dynamic method for multi-task learning in which parameter sharing is conditioned on both the task and the intermediate feature representations of the input instance.
  \item We show empirically that task and instance conditioning enable improved generalization performance across tasks. Experiments on the NYU v2, CityScapes and MIMIC-III datasets demonstrate the potential of the proposed approach, as well as its applicability across problem domains.
\end{itemize}

\section{Related Work}

\textbf{Multi-Task Learning and Parameter Sharing}: Learning multiple tasks with a single model is a challenge that is frequently encountered in natural language processing \cite{Aksoy_2020,liu-etal-2019-multi,Sanh_Wolf_Ruder_2019}, computer vision \cite{liuetal2019mtan,sun2020adashare,vandenhende2020mti}, reinforcement learning \cite{Hessel_2019,Pinto_2016}, and multi-modal learning \cite{Lu_2020_CVPR,Pramanik_2019}, among other areas. Parameter or weight sharing is a well studied mechanism for facilitating shared knowledge transfer in neural networks. For example, in neural architecture search, parameter sharing enables a large number of candidate architectures to be simultaneously trained within a single supernet \cite{guoetal2020,liuetal2019darts,wangetal2021}. Each candidate architecture corresponds to a single execution path in the supernet. In multi-task learning, parameter sharing enables paths for multiple tasks to be simultaneously trained. Traditionally, this has been achieved via so-called hard or soft parameter sharing. Hard parameter sharing \cite{caruana1993,kokkinos2017} involves training a shared backbone of layers followed by task-specific branches. Soft parameter sharing \cite{liuetal2019mtan,ma2018modeling,misraetal2016} involves training multiple task-specific backbones and linking parameters across backbones using regularization or fusion techniques.

Our task conditioning strategy is inspired by the pioneering AdaShare method \cite{sun2020adashare}, 
which revisited traditional parameter sharing patterns in multi-task learning by conditioning layer-wise parameter sharing as a function of the task: a layer can be executed exclusively by one task, shared by multiple tasks, or skipped entirely. This is achieved by learning a task-specific gating policy over the network layers. Each task learns a binary gating vector indicating which layers to execute. 
In contrast, DynaShare conditions parameter sharing on both the task and the intermediate feature representations. Our approach shifts the paradigm presented in AdaShare from ``which layers should tasks share or not share" to include ``how should the task use or not use the layers allocated to it." We show empirically that this additional flexibility in parameter sharing makes it possible to achieve a lift in multi-task accuracy at lower inference cost.

\textbf{Dynamic Neural Networks}: Dynamic neural networks have recently attracted attention in many research communities; a recent survey in the area is presented in \cite{Han2021}. Dynamic neural networks adapt the computation graph of the network depending on the input (instance). Among other applications such as inference acceleration, this adaptivity unlocks the potential of networks that can be trained once and deployed in diverse environments or under different operating conditions. For example, slimmable networks \cite{Sun_2022_CVPR,lietal2021,yuhuang2019,yuetal2019} can be arbitrarily sliced channel-wise at deployment time according to the user's computational resource constraints.

Common strategies for adapting the computation graph in dynamic networks include modifying the input resolution \cite{wangetal2020}, selecting a subset of spatial locations \cite{verelsttuytelaars2020}, selecting a subset of channels \cite{kimetal2018,lietal2021,yuhuang2019,yuetal2019}, or selecting a subset of layers or blocks \cite{veitbelongie2018,wangetal2018,wuetal2018}. DynaShare follows this last strategy and selects a subset of blocks to build a computation graph that depends on both the input and the task. 
Methods for dynamically selecting a subset of layers typically rely on policy networks \cite{Chen_2019,wuetal2018} or gating functions \cite{veitbelongie2018,wangetal2018,Wang_2020}. Policy learning approaches such as BlockDrop \cite{wuetal2018} and GaterNet \cite{Chen_2019} use a policy network that, given an input, produces skipping decisions for all the layers in the backbone network. Gating approaches such as \cite{veitbelongie2018,wangetal2018,Wang_2020} introduce functions responsible for controlling the local inference graph of a layer or block.
DynaShare employs a fully differentiable, hierarchical design consisting of a task-specific policy for coarse layer selection that is further refined by instance-specific gating.

In the context of multi-task learning, Ahn et al. \cite{ahnetal2019} train a dynamic selector subnetwork that determines the level of hierarchical filters in a nested network to activate based on the instance. Our work differs in several ways. In contrast to our instance gates, which use the intermediate feature representations of the previous layers to make dynamic gating decisions, the selector in \cite{ahnetal2019} makes a single-shot level prediction based on the raw network input. The primary motivation of \cite{ahnetal2019} is computational efficiency and model compression, while ours is generalization performance across multiple diverse tasks. Finally, 
DynaShare supports differentiated treatment of tasks, such as balancing different tasks based on difficulty or importance in the operational context. 
We include a comparison with \cite{ahnetal2019} in the NYU v2 and CityScapes experiments.

\section{Proposed Approach}

DynaShare is a flexible parameter sharing method for multi-task learning in which parameter sharing is conditioned on both the task and the intermediate feature representations of the input instance. DynaShare learns from the training data a hierarchical gating policy consisting of a task-specific policy for coarse layer selection and gating units for individual input instances, which work together to determine the execution path at inference time.
In this section, we first provide a high-level overview of the components of the hierarchical gating policy, and then proceed to a more detailed discussion of Gumbel-softmax sampling, the loss functions, and other implementation details.

\subsection{Hierarchical Gating Policy}

DynaShare's task-specific policy establishes a static layer execution policy for each task. The task-specific policy is a discrete vector that determines which layers to execute or skip for each task (Figure \ref{fig:workflow}, top left). Learning a discrete policy in a standard end-to-end differentiable training framework is challenging. We optimize the task-specific policy through the use of Gumbel-Softmax sampling \cite{jang2016categorical}. We will revisit Gumbel-Softmax sampling in the next subsection.
We train the network weights and the parameter sharing policy jointly. To designate a sharing pattern, we sample a task-specific policy from the learned distribution to specify which blocks are selected in different tasks.

DynaShare's instance-specific gating units learn to dynamically adjust the selection decisions of the task-specific policy based on the intermediate feature representations at inference time.
We construct an instance gating method with two parts: estimating relevance of each layer to the input of the same layer, and making the decision to keep or skip the layer based on the estimated relevance \cite{veitbelongie2018}. The relevance estimator is a lightweight design consisting of two convolution layers followed by an activation function. We use the computed output score of the relevance estimator to further make a decision on executing the layer for each individual input. Similar to task-specific policy learning, the discrete controller is trained with the use of Gumbel-Softmax sampling.

\subsection{Gumbel-Softmax Sampling}
\label{gumbelsoftmax}

Given a set of tasks $T=\{\mathcal{T}_1,\mathcal{T}_2,...,\mathcal{T}_K\}$  over a dataset, we use Gumbel-Softmax sampling \cite{jang2016categorical} to learn both task-specific and instance-specific discrete value policies. With this method, instead of sampling from the distribution of a binary random variable to find the optimized policy for each block $\ell_{th}$ and specific task $\mathcal{T}_k$, we can generate the decision from:

\begin{equation}
\label{eq:ulk}
    u_{\ell,k} = \operatorname*{arg\,max}_{j\in \{0,1\}}(\,\log\pi_{\ell,k}(j)+G_{\ell,k}(j)),
\end{equation}

\noindent where $\pi_{\ell,k}=[1-\alpha_{\ell,k},\alpha_{\ell,k}]$ is its distribution vector with $\alpha_{\ell,k}$ representing the probability the $\ell_{th}$ block is executed for task $\mathcal{T}_k$. $G_{\ell,k}=-\log(-\log \, U_{\ell,k})$ are Gumbel random variables with $U_{\ell,k}$ sampled from a uniform distribution. With the reparameterization trick \cite{jang2016categorical}, we can relax the non-differentiable argmax operator in Eq.~\ref{eq:ulk}:

\begin{equation}
\label{eq:ulkrelax}
\small
    u_{\ell,k}(j) = \frac{\exp \, ((\,\log\pi_{\ell,k}(j)+G_{\ell,k}(j))/\tau)}{\sum_{j\in \{0,1\}} \exp \, ((\,\log\pi_{\ell,k}(j)+G_{\ell,k}(j))/\tau)},
\end{equation}

\noindent where $\tau$ is the softmax temperature. When $\tau \to 0$, the softmax function gets closer to the argmax function and we would have a discrete sampler, and when $\tau \to \infty$ it becomes a uniform distribution. Following \cite{sun2020adashare}, for task-specific policy during training we use the relaxed version given by Eq.~\ref{eq:ulkrelax}. 
After training, we sample from the learned distribution to obtain a discrete task-specific policy. For learning the instance-specific policy, we get a discrete sample from Eq.~\ref{eq:ulk} during the forward pass and we compute the gradient of relaxed version given by Eq.~\ref{eq:ulkrelax} in the backward pass.

\subsection{Loss Functions}
\label{subsec:loss_functions}

The goal of DynaShare is to achieve high generalization performance across multiple tasks by employing a flexible parameter sharing strategy. For effective parameter sharing, we need to encourage sharing the blocks among different tasks and prevent the case of splitting parameters with no knowledge sharing. To address these issues, we borrow from AdaShare \cite{sun2020adashare} a  sparsity regularization and sharing loss ($\mathcal{L}_{sparsity}$ and $\mathcal{L}_{sharing}$, respectively) to minimize the log-likelihood of the probability of a block being executed and maximise the knowledge sharing simultaneously:  

\begin{equation}
    \mathcal{L}_{sparsity} = \sum_{\ell\leq L,k\leq K}\log \alpha_{\ell,k},
\end{equation}

\begin{equation}
    \mathcal{L}_{sharing} = \sum_{k_1,k_2 \leq K} \sum_{\ell\leq L}\frac{L-\ell}{L}| \alpha_{\ell,k_1}-\alpha_{\ell,k_2}|,
\end{equation}

\noindent where $L$ is the total number of blocks. For learning the instance gating, we add a loss term that encourages each layer to be executed at a certain target rate $t$. We estimate the execution rates over each mini-batch and penalize deviation from given target rate. We calculate the instance loss as

\begin{equation}
\label{eq:instance}
    \mathcal{L}_{instance} = \sum_{\ell\leq L}(\beta_{\ell} -t)^2,
\end{equation}

\noindent where $\beta_{\ell}$ is the fraction of instances within a mini-batch for which the $\ell_{th}$ block is executed. Including the task-specific losses, the final training loss of DynaShare is:

\begin{align}
    \mathcal{L}_{total} &= \sum_{k}\lambda_k \mathcal{L}_k + \lambda_{sparsity}\mathcal{L}_{sparsity}\nonumber \\  &+ \lambda_{sharing}\mathcal{L}_{sharing} 
    + \lambda_{instance}\mathcal{L}_{instance},
\end{align}

\noindent where $\mathcal{L}_k$ is the task-specific loss, weights $\lambda_k$ are used for task balancing, and $\lambda_{sparsity}$, $\lambda_{sharing}$, $\lambda_{instance}$ are the balance parameters for sparsity, sharing and instance losses, respectively.

\subsection{Task-Instance Fusion: Implementation Details} 
\label{fusion}

\begin{figure}[t]
\begin{center}
\includegraphics[width=\columnwidth]{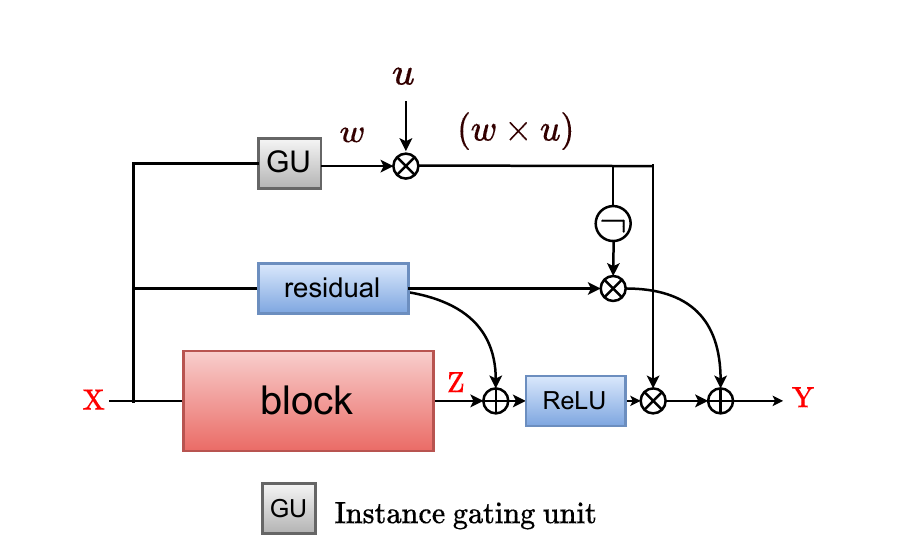}
\end{center}
\caption{Schematic for the fusion of task-specific policy and instance-specific gating outputs. Given the block input $X$, the gating unit GU produces the instance-specific gating output $w \in \{0,1\}$ and the block produces $Z$. The task-specific policy output is denoted by $u \in \{0,1\}$. The resulting gated output $Y$ is is a function of $X$, $Z$, $u$, and $w$ (see text for details).}
\label{fig:fusion_architecture}
\end{figure}

We next provide a full description of how the task-specific policy and instance-specific gating outputs are fused. A schematic is shown in Figure \ref{fig:fusion_architecture}. Given block input $X$, we refine the task-specific policy output $u \in \{0,1\}$ by the instance gating output $w \in \{0,1\}$ to produce block output $Y$ as follows:

\begin{center}
\footnotesize			
$Y = \left\{ \begin{array}{lcl} \text{residual($X$)} & \mbox{if} & \text{$u = 0$ \hspace{0.1cm} or \hspace{0.1cm} $w = 0$} \\
\text{ReLU(residual($X$) + $Z$)} & \mbox{if} & \text{$u = 1$ \hspace{0.1cm} and \hspace{0.1cm}  $w = 1$}
\end{array}\right.
$
\end{center}

\noindent where $Z$ is the output of the convolutional block. In short, when the task-specific policy and instance-specific gating both indicate that the block should be activated ($u=1$ and $w=1$), the resulting block output is $ReLU(residual(X) + Z)$. When either the task-specific policy or instance-specific gating do not activate the block ($u=0$ or $w=0$), the resulting output is $residual(X)$, skipping that block.

\section{Experiments}

In this section, we present experimental results and ablation studies to validate that task and instance conditioning contribute to improved generalization performance in multi-task learning. We compare DynaShare with single-task learning, AdaShare \cite{sun2020adashare}, and other strong multi-task learning baselines. The PyTorch implementation of the code will be made available at {\small\url{https://github.com/BorealisAI/DynaShare}}.  

\subsection{Datasets and  evaluation metrics}

We evaluate the performance of DynaShare on three multi-task learning datasets: NYU v2 \cite{nyuv2}, CityScapes \cite{Cordts2016Cityscapes} and MIMIC-III \cite{harutyunyan2019multitask,johnson2016mimic}. We include datasets from two distinct problem domains to demonstrate the versatility of DynaShare's task and instance conditioned parameter sharing approach.

\textbf{NYU v2} is a commonly adopted benchmark for evaluating multi-task learning methods. We consider the two tasks of predicting semantic segmentation and surface normals. Semantic segmentation is evaluated using mean Intersection over Union (mIoU) and Pixel Accuracy (Pixel acc). For surface normal prediction we report mean and median pixel error, which means the mean and median angle distance between the prediction and ground truth for all pixels. Additionally to mean and median pixel error, we also compute the percentage of predicted pixels within the angles of $11.25^{\circ}$, $22.5^{\circ}$ and $30^{\circ}$.       

\textbf{CityScapes} is another widely used dataset for benchmarking semantic segmentation and depth prediction. Semantic segmentation is evaluated using the same mean IoU and Pixel Accuracy from NYU v2. For depth prediction five metrics are used, absolute and relative errors and measures of the relative difference between prediction and ground truth by the percentage of $\delta = max\left\{ \frac{y_{pred}}{y_{gt}},\frac{y_{gt}}{y_{pred}} \right\}$ within thresholds of $1.25, 1.25^2, 1.25^3$.

\textbf{MIMIC-III} consists of patients information from over 40,000 intensive care units (ICU) stays. The four tasks of this dataset are Phenotype prediction (Pheno), In-hospital mortality prediction (IHM), Length-of-stay (LOS), and Decompensation prediction (Decomp). MIMIC-III is the main benchmark for heterogenous multi-task learning with time series. It is considered heterogenous due to the different task characteristics. The dataset includes two binary tasks, one temporal multi-label task, and one temporal classification task, respectively. 
For a fair comparison we adopted the same split between train, validation, and test set as used by all the previous baselines; the ratios are $70\%$, $15\%$, and $15\%$. The metric for comparing the results is AUC (Area Under The Curve) ROC for the binary tasks and Kappa Score for the multiclass tasks. 

We report the multi-task learning performance of each method with $\Delta$ as defined in \cite{maninis2019attentive}, which compares results with their equivalent single task values:
\begin{equation}
\small
    \Delta_{\mathcal{T}_k} = \frac{1}{|M|}\sum_{j=0}^{|M|}(-1)^{l_j}(M_{\mathcal{T}_k,j} - M_{ST,j})/M_{ST,j}*100\%,
\end{equation}
\noindent where $l_j=1$ if a lower value shows better performance for the metric $M_j$ and $0$ otherwise. $M_{ST,j}$ is the single task value of $j$ metric. We calculate the overall performance by averaging $\Delta_{\mathcal{T}_k}$ over all tasks:
\begin{equation}
    \Delta = \frac{1}{K}\sum_{k \leq K}\Delta_{\mathcal{T}_k},
\end{equation}

\begin{algorithm}[t]
\caption{DynaShare training algorithm}\label{alg:training}
\footnotesize
\centering
\begin{algorithmic}[1]
\State $t \gets 1$
\Comment{target rate equal to 1 means all the layers are executed for all the instances (Eq. \ref{eq:instance})}
\For{warm-up epochs} \Comment{Warm-up stage: hard parameter sharing}
    \State $\mathcal{L}_{Total}=\sum_{k}\lambda_k \mathcal{L}_k$
    \State optimise network weights 
\EndFor
\While{maximum epochs not reached} \Comment{Optimise network and task policy distribution alternatively}
    \For{$e_1$ epochs}
        \State $\mathcal{L}_{Total}=\sum_{k}\lambda_k \mathcal{L}_k$
        \State optimise network weights 
    \EndFor
    \For{$e_2$ epochs}
        \State $\mathcal{L}_{Total}=\sum_{k}\lambda_k \mathcal{L}_k + \lambda_{sparsity}\mathcal{L}_{sparsity} + \lambda_{sharing}\mathcal{L}_{sharing}$
        \State optimise task policy distribution 
    \EndFor
\EndWhile \\
\Return task policy distribution and network weights
\end{algorithmic}
\end{algorithm}

\begin{algorithm}[t]
\caption{DynaShare re-training algorithm}\label{alg:re-training}
\footnotesize
\centering
\begin{algorithmic}[1]
\State $t \gets$ \text{desired target rate}  \Comment{(Eq. \ref{eq:instance})}
\State \text{task network structure} $\gets$ \text{sample of task policy}
\While{maximum epochs not reached}
    \State $\mathcal{L}_{Total}=\sum_{k}\lambda_k \mathcal{L}_k + \lambda_{instance}\mathcal{L}_{instance}$
    \State optimise task network structure with target rate
\EndWhile \\
\Return network weights
\end{algorithmic}
\end{algorithm}

\subsection{Experimental settings}

Following AdaShare's task-specific policy training strategy \cite{sun2020adashare}, for the first few epochs we consider hard parameter sharing by sharing all blocks across tasks. This sets the network at a good starting point for policy learning. Curriculum learning \cite{bengio2009curriculum} is used to encourage a better convergence. The network and task-specific policy distribution parameters are optimized alternatively (Algorithm \ref{alg:training}). After learning the task-specific policy distribution, we sample the distribution and fix the network structure based on task policy. Then, we re-train the network to learn the instance gating using the full training set (Algorithm \ref{alg:re-training}). 

\textbf{NYU v2.} We used the same Deeplab-ResNet-18 backbone as adopted in the experiments of AdaShare \cite{sun2020adashare}. In terms of optimizer we use Adam for task-specific policy learning and SGD for instance gating during the re-training phase. During the re-train, we sample 8 different network architectures from the learned policy and report the best re-train performance as our result (same methodology employed in the AdaShare paper and implemented in their public code). We trained DynaShare for 20000 epochs for training with learning rate to 0.001 that is halved every 10000 epochs, we use cross-entropy for semantic segmentation and inverse of cosine similarity for surface normal prediction. We trained the models using the training set with batch 16 for training and re-training phases. During training we warmup the network (no task policy learning) for 10000 epochs, which is followed by policy learning, with Gumbel-Softmax temperature of 5 decaying at a rate of 0.965 when baseline performance is met or exceeded, for another 10000 epochs. Re-training learns instance gating, with a constant Gumbel-Softmax temperature of 1, for 2000 epochs.

\textbf{Cityscapes.} We used ResNet-34 (16) as backbone, as adopted in the experiments of AdaShare \cite{sun2020adashare}.

\textbf{MIMIC-III.} We used ResNet-18 (8 blocks) as backbone, Adam as optimizer for policy distribution parameters and SGD to update network weights. We set the learning rate to 0.001 that is halved every 250 epochs, binary cross-entropy loss for binary tasks and cross-entropy loss for multi-label tasks. We used the task’s AUC sum in the validation set to define the best model, where the largest sum indicates the best epoch, and consequently, the best model. We trained the models using the training set with batch size of 256 for 1000 epochs.

\subsection{NYU v2 Results}

\begin{table*}[t]
\centering
\small
\begin{tabular}{c|c|c|c|c|cc|ccc|c|c}
\bottomrule[1.2pt]
Method &  & \multicolumn{3}{c|}{$\mathcal{T}_1$: Semantic seg.} & \multicolumn{6}{c|}{$\mathcal{T}_2$: Surface normal prediction} & $\Delta_T \uparrow$ \\ \cline{3-11} 
& \# Params & mIoU $\uparrow$ & Pixel & $\Delta_{T_1} \uparrow$ & \multicolumn{2}{c|}{Error $\downarrow$} & \multicolumn{3}{c|}{$\Delta \theta$, within $\uparrow$} & $\Delta_{T_2} \uparrow$ & \\
& (\%) $\downarrow$ & & Acc $\uparrow$ & & Mean & Median & $11.25^\circ$ & $22.5^\circ$ & $30^\circ$ & & \\
\toprule[0.3pt] \bottomrule[0.3pt]
Single Task & 0.0 & 27.8 & 58.5 & 0.0 & 17.3 & 14.4 & 37.2 & 73.7 & 85.1 & 0.0 & 0.0 \\
Shared Bottom & -50.0 & 22.6 & 55.0 & -12.3 & 16.9 & 13.7 & 41.0 & 73.1 & 84.3 & +3.1 & -4.6\\
DEN \cite{ahnetal2019} & -39.0 & 26.3 & 58.8 & -2.4 & 17.0 & 14.3 & 39.5 & 72.2 & 84.7 & +1.2 & -0.6 \\
Cross-Stitch \cite{misraetal2016} & 0.0 & 25.3 & 57.4 & -5.4 & 16.6 & 13.2 & 43.7 & 72.4 & 83.8 & +5.3 & -0.1\\
Sluice \cite{ruderetal2019} & 0.0 & 26.6 & 59.1 & -1.6 & 16.6 & 13.0 & 44.1 & 73.0 & 83.9 & +6.0 & +2.2 \\
NDRR-CNN \cite{gaoetal2019} & +6.5 & 28.2 & 60.1 & +2.1 & 16.8 & 13.5 & 42.8 & 72.1 & 83.7 & +4.1 & +3.1 \\
DiSparse \cite{Sun_2022_CVPR} & -30.0 & 29.4 & 59.9 & +4.2 & 16.5 & 13.3 & 43.2 & 72.3 & 84.0 & +4.9 & +4.6 \\
MTAN \cite{liuetal2019mtan} & +23.5 & 29.5 & 60.8 & +5.0 & 16.5 & 13.2 & 44.1 & 72.8 & 83.7 & +5.7 & +5.4 \\
AdaShare \cite{sun2020adashare} & -50.0 & 29.6 & 61.3 & +5.6 & 16.6 & 12.9 & 45.0 & 72.1 & 83.2 & +6.2 & +5.9 \\ \hline 
DynaShare (0.55) & -49.1 & \textbf{30.2} & \textbf{61.4} & \textbf{+6.9} & \textbf{13.5} & \textbf{10.3} & \textbf{53.4} & \textbf{80.9} & \textbf{90.3} & \textbf{+21.8} & \textbf{+14.4}  \\
\toprule[1.2pt]
\end{tabular}
\caption{Results on the NYU v2 dataset (semantic segmentation and surface normal prediction).}
\label{tab:nyuv2_results}
\end{table*}




Table \ref{tab:nyuv2_results} shows experimental results on the NYU v2 dataset for the tasks of semantic segmentation and surface normal prediction. Our models are denoted \textit{DynaShare ($t$)}, where $t$ is the target rate (see Eq.~\ref{eq:instance}). The target rate is applied for the layers in the last four ResNet blocks, while the first four ResNet blocks are always enabled. DynaShare outperforms all baselines on 7 out of 7 metrics with an overall performance lift of $\Delta_{T}=+14.4$, which is a relative improvement of $2.4\times$ when compared to the previous state-of-the-art. On the surface normal prediction task, DynaShare more than triples the performance lift of the previous  best performing approach ($3.5\times$ better). 

Table \ref{nyuv2_computation_cost} shows a comparison of the computational cost of the AdaShare and DynaShare models, for multiple instance gating target rates. The table shows  that instance gating adds a small number of parameters ($+1.8\%$ relative) while reducing the FLOPs by $41.8\%$ for the best performing target rate ($0.55$). 
The DynaShare (0.4) model requires half the FLOPs of AdaShare while achieving almost twice the $\Delta_{T}$ improvement.

\begin{table}[t]
\centering
\begin{tabular}{lccc}
\toprule[1.2pt]
& \# Params & FLOPs & $\Delta_T \uparrow$ \\ \hline
AdaShare & 11.19M & 18.48G & +5.9 \\
DynaShare (0.99)& 11.39M & 18.40G & +5.6 \\
DynaShare (0.80)& 11.39M & 15.39G & +11.8 \\
DynaShare (0.67)& 11.39M & 12.99G & +12.3 \\
DynaShare (0.55)& 11.39M & 10.75G & \textbf{+14.4} \\
DynaShare (0.40)& 11.39M & 8.97G & +11.6 \\
\bottomrule[1.2pt]
\end{tabular}
\caption{Comparison of computational cost of DynaShare and AdaShare models on the NYU v2 dataset.}
\label{nyuv2_computation_cost}
\end{table}

\subsection{CityScapes Results}

Table \ref{tab:cityscapes_results} shows experimental results on the Cityscapes 2-Task dataset for the tasks of semantic segmentation and depth prediction. Our models are denoted \textit{DynaShare ($t$)}, where $t$ is the target rate (see Eq.~\ref{eq:instance}). The target rate is applied for the layers in the last nine blocks, while the first seven blocks are always enabled. DynaShare outperforms all baselines with an overall performance lift of $\Delta_{T}=+5.0$, which is a relative improvement of $78\%$ compared to the previous state-of-the-art. On semantic segmentation DynaShare more than doubles the performance lift of the previous best performing approach ($2.3\times$ better). On depth estimation our performance is on par with the previous state-of-the-art. DynaShare also presents the best performance balance among all compared methods. On one hand, Cross-Stitch \cite{misraetal2016} is the best depth estimation method but presents a poor performance on semantic segmentation. On the other hand, DynaShare is the best semantic segmentation technique and also performs on par with the top depth estimation methods. Our approach is the sole method to present high performance on both tasks.       

Table \ref{tab:cityscapes_computation_cost} shows a comparison of the computational cost of the AdaShare and DynaShare models for the CityScapes dataset. The table shows that instance gating adds a small number of parameters ($+1.1\%$ relative) while reducing the FLOPs by $13\%$ for the best performing target rate ($0.80$). For this dataset a lower instance gating rate (sparser gating) has a stronger negative impact on performance, and $80\%$ target coverage is the best setting. 
However, the FLOPs can be reduced further, to almost $30\%$ less than AdaShare, while still achieving better $\Delta_{T}$.

\begin{table*}[t]
\centering
\small
\begin{tabular}{c|c|c|c|c|cc|ccc|c|c}
\bottomrule[1.2pt]
Method &  & \multicolumn{3}{c|}{$\mathcal{T}_1$: Semantic seg.} & \multicolumn{6}{c|}{$\mathcal{T}_2$: Depth prediction} & $\Delta_T \uparrow$ \\ \cline{3-11} 
& \# Params & mIoU $\uparrow$ & Pixel & $\Delta_{T_1} \uparrow$ & \multicolumn{2}{c|}{Error $\downarrow$} & \multicolumn{3}{c|}{$\delta$, within $\uparrow$} & $\Delta_{T_2} \uparrow$ & \\
& $\downarrow$ & & Acc $\uparrow$ & & Abs & Rel & $1.25$ & $1.25^2$ & $1.25^3$ & & \\
\toprule[0.3pt] \bottomrule[0.3pt]
Single Task & 2 & 40.2 & 74.7 & 0.0 & 0.017 & 0.33 & 70.3 & 86.3 & 93.3 & 0.0 & 0.0 \\
Multi\_Task & 1 & 37.7 & 73.8 & -3.7 & 0.018 & 0.34 & 72.4 & 88.3 & 94.2 & -0.5 & -2.1 \\
DEN \cite{ahnetal2019} & 1.12 & 38.0 & 74.2 & -3.1 & 0.017 & 0.37 & 72.3 & 87.1 & 93.4 & -1.6 & -2.4 \\
Sluice \cite{ruderetal2019} & 2 & 39.8 & 74.2 & -0.8 & 0.016 & 0.31 & 73.0 & 88.8 & 94.6 & +4.0 & +1.6 \\
NDRR-CNN \cite{gaoetal2019} & 2.07 & 41.5 & 74.2 & +1.3 & 0.017 & 0.31 & 74.0 & 89.3 & 94.8 & +3.3 & +2.3 \\
DiSparse \cite{Sun_2022_CVPR} & 1.4 & 42.4 & 74.7 & +2.8 & 0.016 & 0.34 & 73.3 & 88.5 & 94.4 & +1.8 & +2.3 \\
MTAN \cite{liuetal2019mtan} & 1.59 & 40.8 & 74.3 & +0.5 & \textbf{0.015} & 0.32 & 75.1 & 89.3 & 94.6 & +4.8 & +2.7 \\
Cross-Stitch \cite{misraetal2016} & 2 & 40.3 & 74.3 & -0.1 & \textbf{0.015} & \textbf{0.30} & 74.2 & 89.3 & \textbf{94.9} & \textbf{+5.8} & +2.8\\
AdaShare \cite{sun2020adashare} & 1 & 41.5 & 74.9 & +1.8 & 0.016 & 0.33 & \textbf{75.5} & \textbf{89.9} & \textbf{94.9} & +3.8 & +2.8 \\ \hline 
DynaShare (0.80) & 1.01 & \textbf{45.0} & \textbf{75.3} & \textbf{+6.5} & 0.016 & 0.33 & 74.8 & 89.5 & 94.7 & +3.5 & \textbf{+5.0}  \\
\toprule[1.2pt]
\end{tabular}
\caption{Results on the CityScapes 2-Task dataset (semantic segmentation and depth prediction).}
\label{tab:cityscapes_results}
\end{table*}

\begin{table*}[h]
\begin{center}
\begin{tabular}{lccccc}
\toprule[1.2pt]
\multicolumn{1}{c}{Method} & {Pheno} & {LOS} & {Decomp} & {IHM} & {$\Delta$} \\
\hline
Single Task       & {77.00}           & 45.00            & 91.00           & 86.00            & -                   \\
Shared Bottom     & 73.36           & 30.60           & 94.12           & 82.71           & -9.28\%
      \\
MCW-LSTM \cite{johnson2016mimic}          & \textbf{77.40}  & 45.00           & 90.50           & 87.00           & +0.28\%             \\
MMoE \cite{ma2018modeling}             & 75.09           & 54.48           & {96.20}      & 90.44           & +7.36\%             \\
MMoEEx \cite{aoki2022}     & 72.44           & {63.45}      & 96.82           & 90.73           & {+11.74}\%            \\
AdaShare \cite{sun2020adashare}         & 76.39           & 71.68           & 82.08           & {94.15}           & {+14.54}\%            \\ \hline
DynaShare & 76.50           &  \textbf{71.86}          & 84.98           &  \textbf{94.25}  & \textbf{+15.50}\%    \\
\bottomrule[1.2pt]
\end{tabular}
\caption{AUC results for the MIMIC-III dataset. DynaShare outperforms all other baselines in two out of four individual tasks, and achieves an overall lift in relative performance when considering the full set of tasks.}
\label{tab:mimic}
\end{center}
\end{table*}

\begin{table}[t]
\centering
\begin{tabular}{lccc}
\toprule[1.2pt]
& \# Params & FLOPs & $\Delta_T \uparrow$ \\ \hline
AdaShare & 21.30M & 33.35G & +2.8 \\
DynaShare (0.99)& 21.53M & 33.40G & +4.3 \\
DynaShare (0.80)& 21.53M & 29.08G & \textbf{+5.0} \\
DynaShare (0.67)& 21.53M & 23.87G & +3.0 \\
\bottomrule[1.2pt]
\end{tabular}
\caption{Comparison of computational cost of DynaShare and AdaShare models on the CityScapes dataset.}
\label{tab:cityscapes_computation_cost}
\end{table}

\subsection{MIMIC-III Results}

For this time series based dataset, we compared DynaShare with single-task training (separate networks trained for each task), hard-parameter sharing, channel wise LSTM (MCW-LSTM~\cite{johnson2016mimic}), MMoE \cite{ma2018modeling}, MMoEEx \cite{aoki2022}, and AdaShare. The full set of results on the MIMIC-III dataset is presented in Table~\ref{tab:mimic}. 
AdaShare and DynaShare are the strongest performers on the dataset, demonstrating the versatility of task and instance conditioned weight sharing in the time series domain.
DynaShare performs modestly better than AdaShare on all four tasks. 
Overall, DynaShare achieves the highest performance in two out of the four tasks, and the highest average delta improvement relative to the single-task model ($+15.50\%$).


\subsection{Ablation Studies}

To evaluate the impact of both the task-specific policy and instance-specific gating components, we conducted ablation experiments on the MIMIC-III dataset. First, we trained the network with a task-specific policy only, which is equivalent to reducing the method to AdaShare. With a task-specific policy only, we achieve a $+2.8\%$ lift in performance compared to the best previous baseline (MMoEEx). Next, we trained the network with instance-specific gating only. This achieves similar performance with state of the art technique. The full results of these ablation studies are reported in Table~\ref{tab:ablation}. In summary, both the task-specific policy and instance-specific gating improve the performance of DynaShare. For ResNet-18 the instance gating function only adds $0.4\%$ to the task-specific network parameters at training, while it improves the average relative performance of all tasks by $ 0.96\%$.

\begin{table*}[t]
\small
\begin{center}
\begin{tabular}{lccccc}
\toprule[1.2pt]
\multicolumn{1}{c}{{Method}} & {Pheno} & {LOS} & {Decomp} & {IHM} & {$\Delta$} \\
\hline
Single Task       & \textbf{77.00}           & 45.00            & \textbf{91.00}           & 86.00            & -                   \\
Task-specific policy          & 76.39           & 71.68           & 82.08           & {94.15}           & {+14.54}\%            \\
Instance-specific gating  & 73.06 & 68.78 & 82.24 &92.10 & +11.30\% \\
Full model (DynaShare) & {76.50}       &  \textbf{71.86}          & {84.98}           &  \textbf{94.25}  & \textbf{+15.50}\%   \\
\bottomrule[1.2pt]
\end{tabular}
\caption{Ablation studies to evaluate the impact of the task-specific policy and instance-specific gating on overall multi-task learning performance on the MIMIC-III dataset.}
\label{tab:ablation}
\end{center}
\end{table*}

\subsection{Limitations} The primary motivation of DynaShare is to boost the generalization performance of multi-task networks via more flexible weight sharing. Model compression is beyond the scope of this paper. Although we include the number of model parameters and FLOPs following previous work on NYU v2 and CityScapes for a fair comparison, these proxy measures of network complexity are understood to be inadequate in predicting device throughout or energy consumption \cite{wangetal2019,yangetal2017}. A rigorous analysis with respect to resource efficiency would require deeper study incorporating device characteristics and the capacity to leverage modern accelerators.

\section{Conclusion}

In this paper, we proposed a flexible task and instance conditioned parameter sharing approach for boosting the generalization performance of multi-task networks. DynaShare conditions layer-wise parameter sharing as a function of both the task and the intermediate feature representations by learning a hierarchical gating policy. Experiments on NYU v2, CityScapes, and MIMIC-III datasets demonstrate the effectiveness of joint task and instance conditioning as well as its versatility across distinct problem domains.

DynaShare's core idea of conditioning on intermediate feature representations to achieve more flexible parameter sharing is potentially applicable to other learning problems in which parameter sharing plays an important role, such as neural architecture search, continual (lifelong) learning, and multimodal learning. We leave these explorations for future work.

{\small
\bibliographystyle{ieee_fullname}
\bibliography{egbib}
}

\end{document}